\documentclass{article}
\usepackage[T1]{fontenc}
\usepackage{spconf,amsmath,amssymb,graphicx,booktabs,algorithm,algpseudocode,hyperref,xcolor}
\usepackage[font=small]{caption}
\usepackage{array}
\hypersetup{hidelinks}
\usepackage{dblfloatfix}

\vspace{-5mm}
\title{LinePilot Digitizer: Line-Plot Recovery\\with Manual and Automatic Calibration}
\vspace{-5mm}
\name{\shortstack{Fengbo Ma\textsuperscript{1,\textdagger}, Rayan Akhtar\textsuperscript{1,\textdagger}, Aakash Joshi\textsuperscript{2,\textdagger}, Xiaoting Li\textsuperscript{1} \\
Haijian Sun\textsuperscript{1}, Zhen Xiang\textsuperscript{1,*}, Xianyan Chen\textsuperscript{1,*}, Yiping Zhao\textsuperscript{1,*}}\thanks{\textsuperscript{\textdagger}Equal contribution. \textsuperscript{*}Corresponding authors: \mbox{\href{mailto:zxiangaa@uga.edu}{zxiangaa@uga.edu}}, \mbox{\href{mailto:xychen@uga.edu}{xychen@uga.edu}}, and \mbox{\href{mailto:zhaoy@uga.edu}{zhaoy@uga.edu}}.}}
\vspace{-5mm}
\address{\textsuperscript{1}University of Georgia \qquad \textsuperscript{2}University of Pennsylvania}
\vspace{-5mm}
\begin{document}
\maketitle
\begin{figure*}[t]
    \centering
    \includegraphics[width=0.89\textwidth]{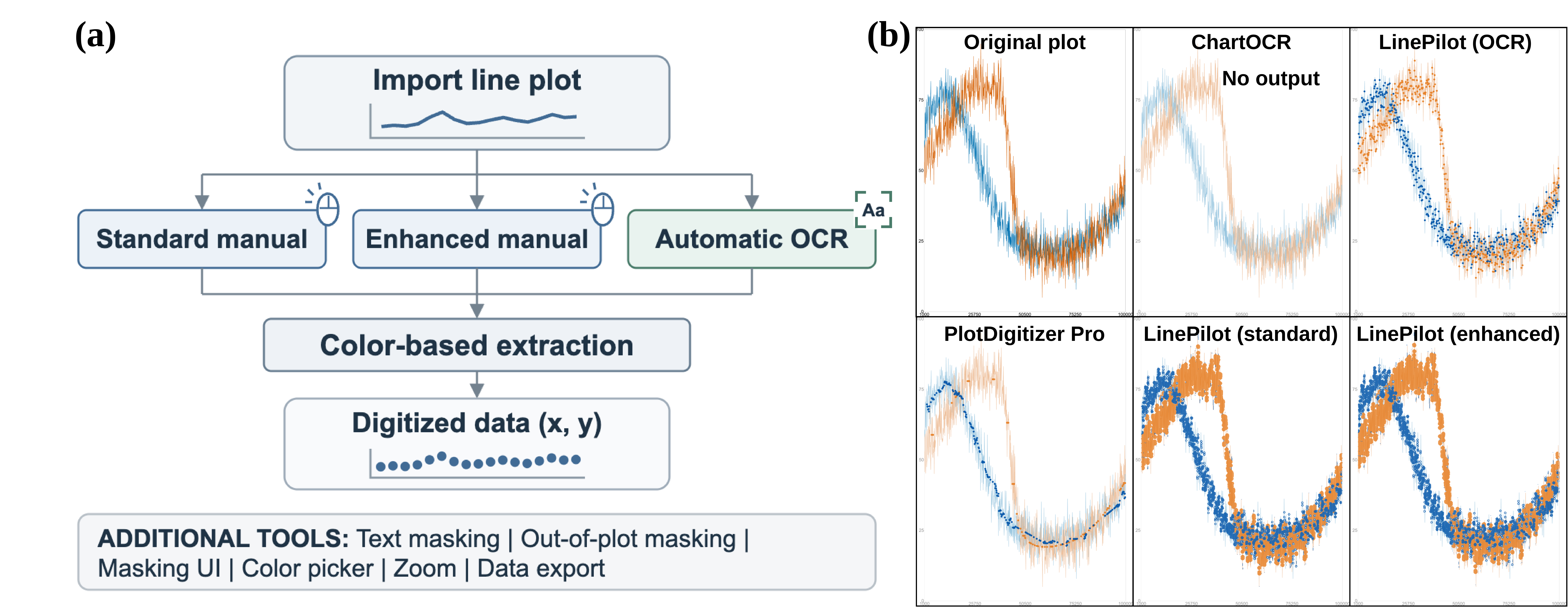}
    \vspace{-3mm}
    \caption{LinePilot workflow and representative digitization results. (a) LinePilot workflow; UI denotes user interface. (b) Digitization results for the \texttt{D0001-overlapping-altair} example from DigitizerBench. Top row: the original plot, the state-of-the-art automatic digitizer ChartOCR~\cite{luo2021chartocr}, which produces no output, and LinePilot (OCR). Bottom row: the state-of-the-art manual digitizer PlotDigitizer Pro~\cite{plotdigitizer}, LinePilot (standard), and LinePilot (enhanced). See the online appendix for results from all methods on this example~\cite{linepilot_supplement}.}
    \label{fig:main}
    \vspace{-3.5mm}
\end{figure*}
\vspace{-5mm}
\begin{abstract}
\vspace{-2mm}
Recovering numerical series from line plots requires accurate axis calibration and reliable curve extraction. We present LinePilot Digitizer (LinePilot), which combines continuous color-based curve recovery with three calibration modes: LinePilot (standard), LinePilot (enhanced), and LinePilot (OCR). We also introduce DigitizerBench, the first dedicated benchmark for systematically evaluating digitizer performance, using an orthogonal design spanning signal, rendering, and plot-structure factors with complementary automatic and human-guided evaluations. We evaluate performance using failure-penalized capped normalized root-mean-square error (FPC-NRMSE), which assigns unit loss to missing, unusable, or catastrophically inaccurate outputs. On DigitizerBench-Full, LinePilot (OCR) achieves the lowest mean FPC-NRMSE (0.672) and highest trusted usability (38.2\%) among the tested automatic pipelines. On DigitizerBench-Lite, LinePilot (enhanced) achieves the lowest mean FPC-NRMSE (0.081), 100\% output success, and highest trusted usability (93.3\%). The orthogonal benchmark design further enables factor analysis to identify the factors that most significantly affect digitizer performance.
Together, the three calibration modes provide a practical trade-off between automation, user control, and accuracy within a shared curve-recovery workflow.
\end{abstract}
\vspace{-5pt}

\vspace{-15pt}
\section{Introduction}
\label{sec:introduction}
\vspace{-10pt}
Charts communicate quantitative evidence in scientific publications~\cite{masry2022chartqa}. Line plots, a widely used type of chart, present continuous measurements and local features in fields such as spectroscopy~\cite{ma2026comprehensive}, mechanical vibration analysis~\cite{smith2015bearing}, and electrocardiography~\cite{baydoun2019ecg}.
Their underlying numerical series support quantitative comparison, reanalysis, data aggregation, machine learning, and literature-analysis systems such as IntrAgent~\cite{ma-etal-2026-intragent}. Yet many publications provide only rendered figures, and source measurements can be difficult to obtain from the authors, leaving the data inaccessible for reanalysis~\cite{jiang2022plot2spectra,baydoun2019ecg}. Consequently, a substantial body of quantitative scientific evidence remains visually accessible to human readers but difficult to reuse computationally. Converting plotted information into machine-readable numerical series therefore provides an important bridge between the published scientific record and modern computational workflows.

Plot digitizers address this need by reconstructing numerical series from rendered line plots, but variations in plot design and rendering make accurate digitization difficult, and errors can affect downstream analysis.
For example, shifted spectral peaks can alter reference matching and chemical interpretation~\cite{hutsebaut2005calibration}, while missed extrema or distorted timing can obscure behavior in vibration and ECG signals~\cite{smith2015bearing,baydoun2019ecg}. Reliable digitization requires accurate pixel-to-axis calibration and complete curve recovery. However, signal noise and morphology, overlapping lines, limited figure width, low resolution, blur, and varying rendering styles complicate both operations, so even a nonempty output may contain miscalibrated coordinates, missing lines, or distorted features.

Existing digitizers cannot address all these challenges through either interactive or automatic workflows. Interactive tools such as PlotDigitizer Pro~\cite{plotdigitizer}, WebPlotDigitizer~\cite{rohatgi_webplotdigitizer}, and Engauge Digitizer~\cite{mitchell_engauge} require users to select axis anchors, curves, or individual points. These operations limit throughput and introduce operator variation, particularly when ticks are small, lines overlap, or images have low resolution. Automatic methods such as ChartOCR~\cite{luo2021chartocr} and LineEX~\cite{shivasankaran2023lineex} reduce human input, but errors in their sequential recognition, calibration, series-grouping, and tracing stages can propagate into sparse, incomplete, or miscalibrated outputs. Existing benchmarks mainly evaluate component detection, data extraction, or question answering~\cite{luo2021chartocr,methani2020plotqa,masry2022chartqa}, leaving end-to-end digitizer reliability under controlled signal and rendering variations insufficiently characterized.


We propose LinePilot, a line-plot digitizer with two stages: pixel-to-data calibration and continuous curve recovery. The calibration variants are LinePilot (standard), LinePilot (enhanced), and LinePilot (OCR); the OCR variant combines recognition results and rejects inconsistent readings. Curve recovery uses dynamic programming to link color-consistent candidates across image columns, rewarding support while penalizing jumps and gaps. We introduce DigitizerBench, which evaluates single, overlapping, and stacked line plots under orthogonally balanced signal and rendering factors. On DigitizerBench-Full, LinePilot (OCR) reduces mean FPC-NRMSE by 0.281 relative to the strongest tested baseline; on DigitizerBench-Lite, LinePilot (enhanced) achieves the lowest mean FPC-NRMSE (0.081) and highest trusted usability (93.3\%).
Contributions are as follows: 1) LinePilot combines continuous color-based curve recovery with three calibration modes---LinePilot (standard), LinePilot (enhanced), and LinePilot (OCR)---balancing user control, automation, and accuracy; 2) DigitizerBench uses an orthogonally balanced design spanning signal, rendering, and plot-structure factors, with DigitizerBench-Full for automatic evaluation and DigitizerBench-Lite for human-guided evaluation; and 3) automatic and human-guided evaluations assess end-to-end reliability and reveal the benchmark factors most affecting performance.

\vspace{-7mm}
\section{Related Work}
\label{sec:related}
\vspace{-10pt}
\noindent\textbf{Interactive plot digitizers.} PlotDigitizer Pro~\cite{plotdigitizer} provides manual axis anchors, point selection, and zoom; WebPlotDigitizer~\cite{rohatgi_webplotdigitizer} combines manual calibration with computer-vision-assisted extraction; and Engauge Digitizer~\cite{mitchell_engauge} supports axis-point calibration and assisted curve tracing. Our human-guided evaluation compares these workflows with LinePilot (standard) and LinePilot (enhanced) to assess calibration accuracy and recovery quality.

\noindent\textbf{Automatic chart and line extraction.} ChartOCR~\cite{luo2021chartocr} combines learned keypoint detection with chart-specific extraction rules. LineEX~\cite{shivasankaran2023lineex} integrates keypoint extraction, text recognition, and series grouping and scaling. CHART-Info~\cite{chartinfo2020} evaluates axis analysis, plot-element detection, and numerical data recovery, including an end-to-end image-only task. These approaches motivate separate evaluation of calibration, series completeness, and curve accuracy.

\noindent\textbf{Benchmarks for chart understanding.} ExcelChart400K~\cite{luo2021chartocr} supports chart-component detection and numerical data extraction. PlotQA~\cite{methani2020plotqa} tests numerical reasoning through question answering (QA) over scientific plots. ChartQA~\cite{masry2022chartqa} emphasizes visual and logical reasoning with human-written and generated questions. These benchmarks primarily target chart-data extraction and QA. Among the benchmarks reviewed, none couples an orthogonally balanced design of signal and rendering conditions with systematic digitizer evaluation, the focus of DigitizerBench (Table~\ref{tab:benchmarks}).

\begin{table}[t]
\centering\small
\caption{Verified line-layout coverage and experimental design ($\checkmark$: covered; $\times$: not covered).}
\vspace{-10pt}
\label{tab:benchmarks}
\setlength{\tabcolsep}{3pt}
\begin{tabular}{@{}>{\raggedright\arraybackslash}p{0.31\columnwidth}>{\centering\arraybackslash}p{0.23\columnwidth}>{\centering\arraybackslash}p{0.20\columnwidth}>{\centering\arraybackslash}p{0.18\columnwidth}@{}}
\toprule
Benchmark & Line types covered & Controlled factors & Balanced design \\
\midrule
ExcelChart400K~\cite{luo2021chartocr} & 1 & $\times$ & $\times$ \\
PlotQA~\cite{methani2020plotqa} & 2 & $\checkmark$ & $\times$ \\
ChartQA~\cite{masry2022chartqa} & 2 & $\times$ & $\times$ \\
\textbf{DigitizerBench} & \textbf{3} & $\boldsymbol{\checkmark}$ & $\boldsymbol{\checkmark}$ \\
\bottomrule
\end{tabular}
\vspace{-6mm}
\end{table}

\vspace{-5mm}
\section{Methodology}
\label{sec:methods}
\vspace{-10pt}
LinePilot separates digitization into pixel-to-data calibration and continuous curve recovery. The design follows two observations: calibration errors propagate to every recovered point, and a plotted line forms a continuous path rather than independent pixels. LinePilot (standard), LinePilot (enhanced), and LinePilot (OCR) therefore share the same recovery procedure (Figure~\ref{fig:main}(a)), so the calibration mechanism changes without changing curve extraction.

\vspace{-10pt}
\subsection{Calibration and coordinate mapping}
\vspace{-5pt}
LinePilot (standard) uses the operator's calibration points directly. LinePilot (enhanced) searches near each click for the intended tick or axis stroke and estimates its center from local image intensity. A detected crossing refines both coordinates, whereas a single stroke refines only the coordinate across it. If no suitable stroke is found, or the proposed movement is too large, the original click is retained. LinePilot (enhanced) therefore adds precision only where the image provides supporting evidence.

LinePilot (OCR) first identifies the plotting region and crops the two axis neighborhoods. It reads numerical labels with Tesseract~\cite{smith2007tesseract} page-segmentation modes 6 and 11 and, when available, RapidOCR~\cite{rapidocr2021}. Readings at nearby axis positions are merged, and inconsistent readings are rejected when the remaining ticks support a common scale. Horizontal-axis labels are associated with ticks above them, while vertical-axis labels are associated with ticks to their right. The extreme consistent label--tick pairs define the anchors. LinePilot (OCR) reports failure when all required endpoints cannot be established; manual correction remains available in the interactive workflow but is excluded from automatic evaluation.

For either linear axis, anchors $(p_1,q_1)$ and $(p_2,q_2)$ define $q=q_1+(p-p_1)(q_2-q_1)/(p_2-p_1)$, with $p_1\ne p_2$. We set $(p,q)=(u,x)$ horizontally and $(p,q)=(v,y)$ vertically, where $(u,v)$ are pixel coordinates and $(x,y)$ are data coordinates. The signed denominator accounts for image coordinates increasing downward.

\vspace{-6mm}
\subsection{Curve extraction}
\vspace{-3mm}
After calibration, LinePilot recovers each target curve as a continuous path through color-matched image evidence. Target colors are detected automatically or selected by the operator. Within the plotting region, pixels within tolerance $\tau$ of a target color are grouped into vertical runs in each column. Each run midpoint is a candidate centerline position, and its thickness supplies additional stroke evidence.

Dynamic programming links candidates across columns by rewarding sustained support and stroke thickness while penalizing abrupt vertical movement and gaps. Joint path selection reduces interference from isolated color-matched objects that lack support in neighboring columns. Selected midpoints are transformed to data coordinates by the calibrated mapping. Only image-supported points are exported; missing columns are not filled with synthesized observations.
\setlength{\textfloatsep}{4pt plus 1pt minus 1pt}
\setlength{\intextsep}{4pt plus 1pt minus 1pt}
\begin{algorithm}
\small
\caption{Color-based curve extraction}
\label{alg:column-scanning}
\begin{algorithmic}[1]
\Require Image $I$, region $R$, tolerance $\tau$, mapping $A$
\Ensure Calibrated series $\mathcal S$
\State $\mathcal C\gets$ detected or user-selected colors
\State $\mathcal S\gets\emptyset$
\For{each color $c\in\mathcal C$}
    \State $\mathcal V\gets\emptyset$
    \For{each column $u$ in $R$}
        \State Group pixels within $\tau$ of $c$ into runs
        \State Add run midpoints and thicknesses to $\mathcal V$
    \EndFor
    \State $P\gets$ dynamic-programming path through $\mathcal V$
    \Statex \hspace{\algorithmicindent}Reward support and thickness; penalize jumps and gaps
    \If{$P\ne\emptyset$}
        \State $\mathcal S\gets\mathcal S\cup\{\{A(u,v):(u,v)\in P\}\}$
    \EndIf
\EndFor
\State Export $\mathcal S$ and inspect its overlay on $I$
\end{algorithmic}
\end{algorithm}

The procedure selects one path per target color. Weak color separation, low resolution, and overlapping curves with the same color can therefore prevent complete recovery. Sampling at image-column resolution also cannot restore features unresolved in the source image.

\vspace{-5mm}
\section{Experimental Design}
\label{sec:design}
\vspace{-3.5mm}
\subsection{DigitizerBench construction}
\vspace{-3mm}
An exhaustive factorial design of the 18 five-level signal and rendering factors requires $5^{18}$ conditions. For tractable, balanced coverage, DigitizerBench uses four complementary strength-3 $\mathrm{OA}(625,19,5,3)$ panels~\cite{hedayat1999orthogonal}. Each panel assigns levels to the 18 factors and one generation block, balancing all level combinations across any three design columns. Together, the panels provide a 2,500-image orthogonal-array core for estimating main effects with reduced confounding.

For each configuration, independent traces are synthesized by combining a baseline trend, controlled peaks and valleys, and Gaussian signal noise according to Table~\ref{tab:factors}. The traces form single, overlapping, or vertically stacked plots. Stacked traces are offset to prevent intersection, whereas overlapping traces share the same axes without forced crossings. After joint mapping to the assigned axis ranges, each configuration is rendered with Matplotlib, Plotly, or Altair. Exact curve and tick coordinates are stored as ground truth, while digitizers receive only the rendered images.

To connect the automatic and human evaluations, a mixed-integer linear programming (MILP) selection~\cite{gleixner2021miplib} identifies 60 base configurations balanced across panels, plot types, renderers, and colors. Each is rendered under all nine plot-type--renderer combinations, adding 480 images to the 2,500-image orthogonal-array core for a total of 2,980 images in DigitizerBench-Full. The 60 core images form DigitizerBench-Lite, while all 540 versions form a matched bridge subset within DigitizerBench-Full. For replication, the benchmark package records the master seed, factor levels, orthogonal-array matrices, trace-generation rules, bridge identifiers, software versions, and image manifest. Construction details are provided in the online appendix~\cite{linepilot_supplement}.


\begin{table}
\vspace{-5mm}
\centering\small
\caption{Signal, rendering, and structural factors in DigitizerBench.}
\vspace{-10pt}
\label{tab:factors}
\setlength{\tabcolsep}{4pt}
\begin{tabular}{@{}p{0.20\columnwidth}p{0.59\columnwidth}p{0.13\columnwidth}@{}}
\toprule
Group & Factors & Levels \\
\midrule
Signal (8) & Baseline family and magnitude; extrema count and prominence; feature width; sharpness; asymmetry; signal noise. & 5 each \\
Rendering (10) & Thickness; color; DPI; blur; figure width and aspect ratio; tick-label size; tick count; x- and y-axis ranges. & 5 each \\
Plot type & Single, overlapping, stacked. & 3 \\
Line count & 2, 3, or 5 for multiline plots; 1 for single-line plots. & 3\textsuperscript{*} \\
Renderer & Matplotlib, Plotly, Altair. & 3 \\
\bottomrule
\end{tabular}
\vspace{-2pt}
\par\smallskip\small\textsuperscript{*}Conditional on a multiline plot. Example five-level settings: line thickness: 1, 2, 3, 4, and 5 pixels; DPI: 72, 100, 150, 300, and 600.
\vspace{-2mm}
\end{table}

\vspace{-2.5mm}
\subsection{Evaluation protocol}
\vspace{-3mm}
Using released code, we evaluate ChartOCR~\cite{luo2021chartocr}, LineEX~\cite{shivasankaran2023lineex}, and LinePilot (OCR) on all DigitizerBench-Full images. Every DigitizerBench-Lite image is evaluated with PlotDigitizer Pro~\cite{plotdigitizer}, WebPlotDigitizer~\cite{rohatgi_webplotdigitizer}, Engauge Digitizer~\cite{mitchell_engauge}, LinePilot (standard), and LinePilot (enhanced).

\vspace{-6mm}
\subsection{Evaluation metrics}
\label{sec:metrics}
\vspace{-3mm}
\textit{Failure-penalized capped normalized root-mean-square error (FPC-NRMSE)} is the primary metric. For each candidate line match, prediction and ground truth are interpolated without extrapolation at one position per horizontal plot pixel in their shared x-range. Hungarian assignment minimizes $\min(e_{i\ell},1)+(1-C_{i\ell})$. For image $i$, $T_i=1$ only when the predicted line count is correct and every match has finite error, at least 20 valid coordinate pairs, and coverage $C_{i\ell}\geq0.800$:
\vspace{-2mm}
{\small
\begin{equation}
\begin{aligned}
e_{i\ell}
&=\frac{\sqrt{K_{i\ell}^{-1}\sum_{j=1}^{K_{i\ell}}
(\hat y_{i\ell j}-y_{i\ell j})^2}}
{y_{i,\max}-y_{i,\min}},
\qquad
\bar e_i=G_i^{-1}\sum_{\ell=1}^{G_i}e_{i\ell},
\\[-1mm]
F_i
&=
\begin{cases}
\bar e_i, & T_i=1\ \text{and}\ \bar e_i<1,\\
1, & \text{otherwise},
\end{cases}
\qquad
\overline F=N^{-1}\sum_{i=1}^{N}F_i.
\end{aligned}
\label{eq:fpc-nrmse}
\end{equation}
\vspace{-5mm}
}

Here, $K_{i\ell}$ is the evaluated grid size, $G_i$ is the ground-truth line count, $C_{i\ell}=K_{i\ell}/W_i$, and $W_i$ is the endpoint-tick plot width. Thus $\overline F$ equally weights all $N$ images and assigns unit loss to absent, structurally unusable, or full-span-error outputs; lower is better.

\textit{Output success (OS)} is the percentage of images producing valid, nonempty output. \textit{Trusted usability (TU)} is the percentage satisfying $T_i=1$. \textit{Digitized points per line (DPPL)} averages points first across predicted lines within each successful image and then across successful images.

For LinePilot (OCR), a main-effects analysis of variance (ANOVA) uses $F_i$ for all DigitizerBench-Full images, with panel, generation block, and image role as nuisance terms; conditional factors use their active populations. Adjusted sums of squares are used to describe relative effect magnitude. Six prespecified two-factor interactions are examined in separate exploratory models by adding each interaction to the main-effects model. We report $F$ tests, unadjusted $p$ values, and adjusted-sum-of-squares contribution; clustering by base image provides a sensitivity analysis for bridge renderings.

\vspace{-10pt}
\begin{table}
\centering\small
\caption{Automatic results on DigitizerBench-Full and human-guided results on DigitizerBench-Lite. Metrics are failure-penalized capped normalized root-mean-square error (FPC-NRMSE), output success (OS), trusted usability (TU), and digitized points per line (DPPL); arrows indicate preferred directions.}
\vspace{-10pt}
\label{tab:automatic}
\setlength{\tabcolsep}{1.2pt}
\renewcommand{\arraystretch}{0.95}
\begin{tabular}{@{}lrrrr@{}}
\toprule
Method & FPC-NRMSE $\downarrow$ & OS (\%) $\uparrow$ & TU (\%) $\uparrow$ & DPPL $\uparrow$\\
\midrule
\multicolumn{5}{l}{\textit{Automatic: DigitizerBench-Full (2,980 images/method)}}\\
ChartOCR & 0.953 & 30.2 & 5.37 & 29.2 \\
LineEX & 0.997 & 36.6 & 0.369 & 13.5 \\
\textbf{LinePilot (OCR)} & \textbf{0.672} & \textbf{55.1} & \textbf{38.2} & \textbf{1,180} \\
\midrule
\multicolumn{5}{l}{\textit{Human-guided: DigitizerBench-Lite (60 images/method)}}\\
PlotDigitizer Pro & 0.097 & \textbf{100.0} & 91.7 & 474.7 \\
WebPlotDigitizer & 0.164 & 98.3 & 86.7 & 194.6 \\
Engauge Digitizer & 0.790 & 55.0 & 23.3 & 235.3 \\
\textbf{LinePilot (standard)} & 0.101 & \textbf{100.0} & 91.7 & 1024.8 \\
\textbf{LinePilot (enhanced)} & \textbf{0.081} & \textbf{100.0} & \textbf{93.3} & \textbf{1026.5} \\
\bottomrule
\end{tabular}
\vspace{-2mm}
\end{table}

\vspace{-2mm}
\section{Results and Discussion}
\label{sec:results}
\vspace{-3.5mm}
\subsection{Automatic and human-guided performance}
\vspace{-2.3mm}
On DigitizerBench-Full, LinePilot (OCR) performs best automatically: its mean FPC-NRMSE is 0.672 versus 0.953 for ChartOCR and 0.997 for LineEX, while leading in output success (OS) and trusted usability (TU) (Table~\ref{tab:automatic}). Because FPC-NRMSE retains failed and unusable cases, it captures reliability differences excluded by success-conditioned errors.
On DigitizerBench-Lite, LinePilot (enhanced) achieves the strongest human-guided performance, with the lowest mean FPC-NRMSE (0.081), 100\% OS, 93.3\% TU, and the highest DPPL (1026.5). PlotDigitizer Pro and LinePilot (standard) follow at 0.097 and 0.101, while WebPlotDigitizer (0.164) and Engauge Digitizer (0.790) show larger errors. Overall, LinePilot provides competitive performance across automatic and human-guided settings, with LinePilot (enhanced) offering the best balance of accuracy, reliability, and sampling density under the tested conditions.

\vspace{-5mm}
\subsection{Effects of benchmark factors}
\vspace{-2.5mm}
The ANOVA $p$ values for DigitizerBench-Full show that LinePilot (OCR) is affected primarily by image presentation rather than signal complexity. Figure width has the largest effect, followed by DPI and renderer; axis range, tick density, plot type, and blur have significant effects, while signal noise and most curve-shape factors are not significant. The significant width--DPI interaction shows that physical size and raster resolution matter jointly.

For the human-guided evaluation, line thickness is significant ($p<0.05$). One-pixel curves perform poorly, while performance stabilizes across 3--5 pixels. This contrast suggests that human judgment can compensate for lower DPI through better tick reading and reconstruction, whereas very thin strokes provide too little color evidence for reliable line selection and tracing. These findings prioritize resolution-aware preprocessing, robust color extraction, tick-label handling, and calibration validation. See the online appendix for detailed factor results~\cite{linepilot_supplement}.

\vspace{-5mm}
\subsection{Case studies and limitations}
\vspace{-2.5mm}
Figure~\ref{fig:main}(b) compares automatic and human-guided tools on a common DigitizerBench example. Current results cover synthetic plots with linear axes and specified adapters; real-figure accuracy requires attributed examples and reference data. Human results are limited to DigitizerBench-Lite, and inconsistent runtime records prevent uniform speed comparison.

\vspace{-4.5mm}
\section{Conclusion}
\vspace{-3.5mm}
LinePilot unifies standard, enhanced, and OCR calibration with continuous curve recovery, while DigitizerBench evaluates performance under controlled signal and rendering variation. LinePilot (OCR) leads the tested automatic pipelines on DigitizerBench-Full, and LinePilot (enhanced) leads the human-guided methods on DigitizerBench-Lite; factor analysis identifies image size and resolution for automatic recovery and one-pixel line thickness for human-guided recovery as the main vulnerabilities.

\newpage
\section{Acknowledgment}
This work was supported by the National Science Foundation (NSF) through the Pathways to Open-Source Ecosystems (POSE) Program, award No. 2518273. The authors gratefully acknowledge the support and guidance from, and collaboration with, all members of the SpectraGuru\textsuperscript{TM} development and stakeholder teams.

\bibliographystyle{IEEEbib}
\bibliography{strings,refs}
\end{document}